\documentclass[cameraready]{Interspeech}

\title{Vavanagi: a Community-run Platform for Documentation of the Hula Language in Papua New Guinea}

\author[affiliation={1}, orcid=0009-0002-5675-7961, equalcontribution, correspondingauthor]{Bri}{Olewale}
\author[affiliation={2}, orcid=0009-0007-3242-2311, equalcontribution]{Raphael}{Merx}
\author[affiliation={2}, orcid=0000-0002-4058-5459]{Ekaterina}{Vylomova}

\address{
    $^1$ Vula'a Kunenai Community, Central Province, Papua New Guinea \\
    $^2$ The University of Melbourne, Melbourne, Australia
}

\email{hulahistoryproject@gmail.com}


\usepackage{comment}
\usepackage{booktabs}
\usepackage{graphicx}
\usepackage[table]{xcolor}

\definecolor{extbg}{RGB}{242,244,247}
\definecolor{shrbg}{RGB}{230,243,230}
\definecolor{combg}{RGB}{205,233,210}
\newcommand{\extcell}[1]{\cellcolor{extbg}#1}
\newcommand{\shrcell}[1]{\cellcolor{shrbg}#1}
\newcommand{\comcell}[1]{\cellcolor{combg}#1}

\usepackage{tikz}
\usetikzlibrary{shapes.geometric,arrows.meta,positioning,calc}
\tikzset{
    base/.style = {draw, text width=2.8cm, align=center, minimum height=1em, font=\small},
    inout/.style = {base, rectangle, rounded corners},
    process/.style = {base, rectangle, fill=blue!20},
    collab/.style = {base, rectangle, fill=green!20},
}
\newcommand{\workflowstep}[2]{\textbf{\textsc{#1}}\\[-0.15em]\footnotesize #2}

\begin{document}


\maketitle

\begin{abstract}
    We present Vavanagi, a community-run platform for Hula (Vula'a), an Austronesian language of Papua New Guinea with approximately 10,000 speakers. Vavanagi supports crowdsourced English-Hula text translation and voice recording, with elder-led review and community-governed data infrastructure. To date, 77 translators and 4 reviewers have produced over 12k parallel sentence pairs covering 9k unique Hula words. We also propose a multi-level framework for measuring community involvement, from consultation to fully community-initiated and governed projects. We position Vavanagi at Level 5: initiative, design, implementation, and data governance all sit within the Hula community, making it, to our knowledge, the first community-led language technology initiative for a language of this size. Vavanagi shows how language technology can bridge village-based and urban members, connect generations, and support cultural heritage on the community's own terms.
\end{abstract}

\section{Introduction}

Papua New Guinea (PNG) is widely recognized for its high linguistic diversity \cite{foley2000languages,aikhenvaldstebbins2007languages,aikhenvald2014living}. Hula (Vula'a), spoken in Central Province, belongs to the Austronesian language family \cite{ross1988protooceanic,dutton1973checklist}, and has approximately 10,000 speakers \cite{ethnologuehul}. As in many PNG settings, language shifts linked to urban mobility and the dominance of Tok Pisin create pressure on Hula's position, and demand for practical, community-owned language documentation \cite{schreyerwagner2022uncertainty}.

\begin{figure}[t]
    \begin{center}
        \includegraphics[width=0.95\linewidth]{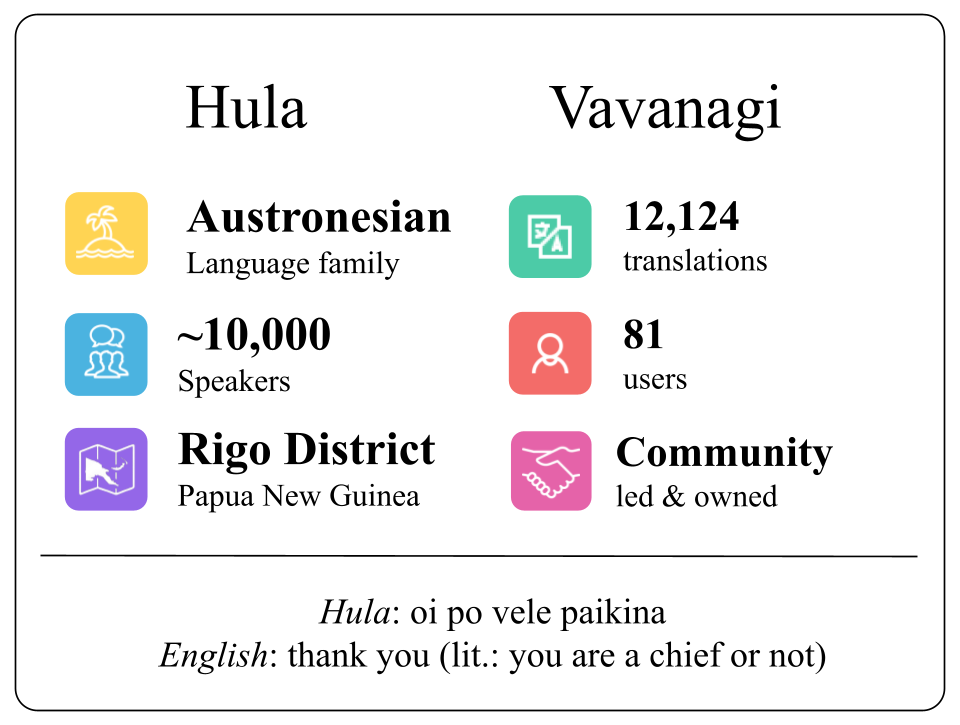}
    \end{center}
    \caption{Overview of the Vavanagi platform and Hula language}
    \label{fig:overview}
\end{figure}

We present Vavanagi (Figure \ref{fig:overview}), a community-run platform for collecting English-Hula parallel text and voice data through crowdsourced translation and elder-led review. The resulting corpus is intended to serve as the foundation for downstream speech technologies, including machine translation (MT) and automated speech recognition (ASR). We document how Vavanagi is the first community-led language technology initiative for a language of this size, with project initiative, design, implementation, and data governance all within the community, demonstrating data sovereignty in practice.

To situate this contribution within the landscape of community-based language technology, we also propose a five-level community involvement spectrum, from community consultation to community-governed work (Figure \ref{fig:community-spectrum}). This spectrum is intended as a reusable conceptual tool for the field, providing shared vocabulary for comparing the degree of community agency across documentation and technology projects.

\section{Background: Community involvement in language technology}



\subsection{Community involvement in language documentation}

\textbf{In linguistic research.}
Collaborative fieldwork has long been debated in linguistics. Bowern and Warner \cite{bowern2015lone} argue that community agency and research goals are mutually beneficial and that ethical documentation requires respecting community sovereignty. More recent frameworks go further, distinguishing gradations from community-based to community-led research in which design authority progressively shifts to the community \cite{bax2024milpa}. Evidence from Pacific contexts shows that community-led documentation can yield larger and higher-quality datasets than visiting-linguist approaches, while supporting long-term sustainability of the work \cite{krajinovic2022communityled}.

\textbf{In language technology.}
Participatory approaches in text and speech technology have grown considerably. At scale, the Masakhane initiative demonstrated that grassroots collaboration can produce MT datasets and benchmarks for 30+ African languages \cite{masakhane-mt-emnlp20,adelani-etal-2022-thousand}. Tool design work increasingly centres linguistic sovereignty \cite{oneil-etal-2024-computational}, and position papers argue that serving communities requires prioritising their needs rather than treating them as data sources \cite{liu-etal-2022-always}. Process-focused work extends this to speech technology, advocating participatory design at every stage \cite{hutchinson2025designing,cooper-etal-2024-things}, while platforms such as the Mozilla Data Collective \cite{mdc2025} position contributors as data stewards who govern how their datasets are licensed and used.

\begin{figure*}[t]
    \centering
    \includegraphics[width=\linewidth]{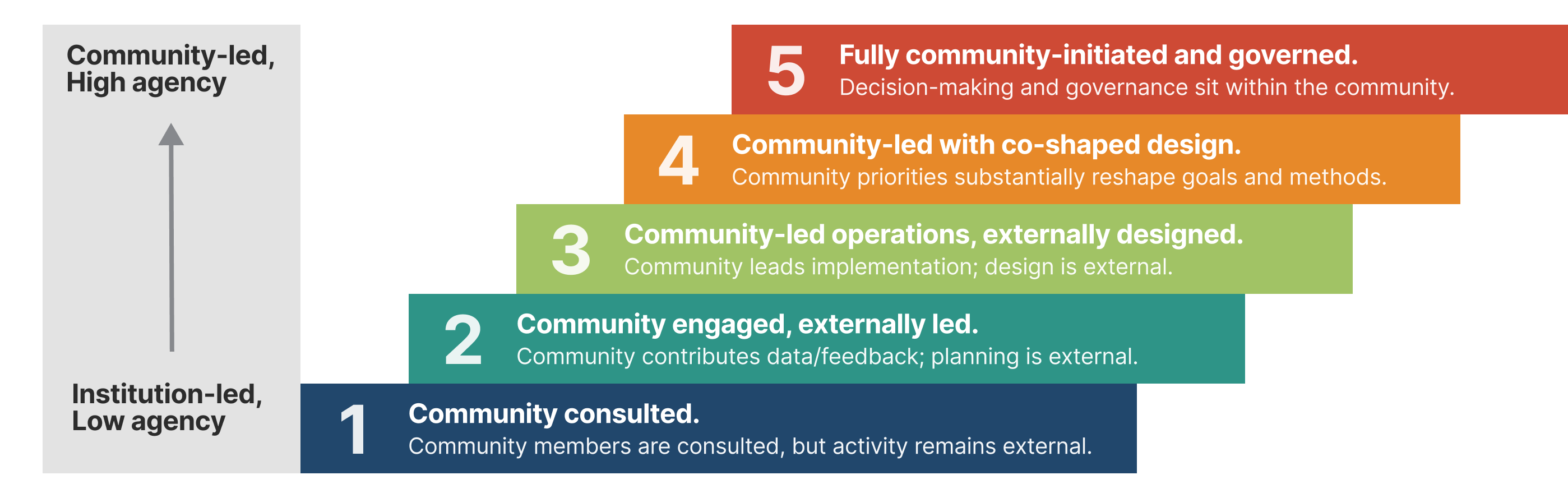}
    \caption{Community involvement spectrum, from consultation to community governance. See mapping of prior work onto the spectrum in Table \ref{tab:involvement-spectrum}.}
    \label{fig:community-spectrum}
\end{figure*}

\subsection{Community involvement spectrum}
Prior work emphasizes participatory practice, but community participation takes meaningfully different forms \cite{watimelu2024community,cox-etal-2025-creating,hutchinson2025designing,oneil-etal-2024-computational}. Without explicit distinctions, projects ranging from occasional consultation to full community governance may all be described as ``community-based'', obscuring the extent of community agency. A shared vocabulary is needed to compare projects and identify where authority truly lies.

Four dimensions capture where a project sits on this continuum: (i)~\textbf{initiative origin}: who starts the project; (ii)~\textbf{design authority}: who shapes its goals and methods; (iii)~\textbf{operational leadership}: who leads day-to-day activity; and (iv)~\textbf{data governance}: who owns and controls the resulting data. These dimensions tend to stack: externally initiated projects are more likely to retain external design authority, which in turn correlates with external operational control. This ordering yields a five-level spectrum (Figure~\ref{fig:community-spectrum}).

\begin{itemize}
    \item \textbf{Level 1: Community consulted.} Community members are consulted (e.g., before project launch), but day-to-day project activity remains external.
    \item \textbf{Level 2: Community engaged, externally led.} Community members contribute data or feedback, while planning and operational control remain with external teams.
    \item \textbf{Level 3: Community-led operations, externally designed.} Community members lead daily implementation, but core project design originates outside the community.
    \item \textbf{Level 4: Community-led with co-shaped design.} Project initiative starts externally, but community priorities substantially reshape goals, methods, and outputs.
    \item \textbf{Level 5: Fully community-initiated and governed.} Initiative, decision-making, implementation, and data governance all sit within the language community.
\end{itemize}

Table~\ref{tab:involvement-spectrum} maps a selection of prior work (corpus gathering for very low-resource languages) onto this spectrum. Community-led initiatives for low-resource languages have become more common in recent years, including in Africa \cite{masakhane-mt-emnlp20}, Central Asia \cite{mirzakhalov2021large}, and the Middle East \cite{hameed25_interspeech}, but these efforts target languages with speaker populations in the millions, at least two orders of magnitude larger than Hula. To our knowledge, Vavanagi is the first community-led language documentation project for a language of this size.

\begin{table}[t]
\centering
\small
\caption{Mapping of prior work onto the community involvement spectrum. Columns: \textbf{I}nitiative origin, \textbf{D}esign authority, \textbf{O}perational leadership, data \textbf{G}overnance. Codes: \extcell{E}\,=\,externally led, \shrcell{S}\,=\,shared/co-shaped, \comcell{C}\,=\,community-led.}
\label{tab:involvement-spectrum}
\begin{tabular}{p{2.8cm} cccc r}
\toprule
\textbf{Work} & \textbf{I} & \textbf{D} & \textbf{O} & \textbf{G} & \textbf{Level} \\
\midrule
Petrariu et al.\ \cite{petrariu-nisioi-2024-multilingual} -- \textit{Aromanian} & \extcell{E} & \extcell{E} & \extcell{E} & \extcell{E} & 2 \\
Li et al.\ \cite{li-etal-2024-speech} -- \textit{Khinalug} & \extcell{E} & \extcell{E} & \extcell{E} & \extcell{E} & 2 \\
Prabowo et al.\ \cite{prabowo2024preserving} -- \textit{Meher/Woirata} & \extcell{E} & \extcell{E} & \comcell{C} & \shrcell{S} & 3 \\
Cox et al.\ \cite{cox-etal-2025-creating} -- \textit{Tsuut'ina} & \extcell{E} & \shrcell{S} & \shrcell{S} & \shrcell{S} & 4 \\
Frontull et al.\ \cite{frontull-etal-2025-bringing} -- \textit{Ladin} & \shrcell{S} & \shrcell{S} & \comcell{C} & \shrcell{S} & 4 \\
Krajinovic et al.\ \cite{krajinovic2022communityled} -- \textit{Nafsan} & \shrcell{S} & \shrcell{S} & \comcell{C} & \shrcell{S} & 4 \\
\textbf{Vavanagi (this work)} -- \textit{Hula} & \comcell{C} & \comcell{C} & \comcell{C} & \comcell{C} & 5 \\
\bottomrule
\end{tabular}
\end{table}

\section{Methodology}

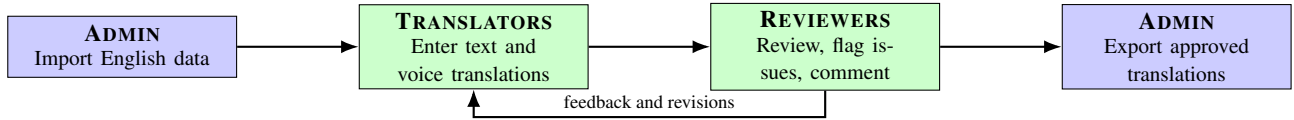
\begin{figure*}[t]
\centering
\begin{tikzpicture}[node distance=1.2cm and 0.9cm, auto]
    \node [process] (adminin) {\workflowstep{Admin}{Import English data}};
    \node [collab, right=1.6cm of adminin] (trans) {\workflowstep{Translators}{Enter text and voice translations}};
    \node [collab, right=1.6cm of trans] (review) {\workflowstep{Reviewers}{Review, flag issues, comment}};
    \node [process, right=1.6cm of review] (adminout) {\workflowstep{Admin}{Export approved translations}};

    \draw[-Latex, thick] (adminin) -- (trans);
    \draw[-Latex, thick] (trans) -- (review);
    \draw[-Latex, thick] (review) -- (adminout);

    \draw[-Latex, thick] (review.south) -- ++(0,-0.35) -| (trans.south);
    \node[above, font=\scriptsize] at ($(trans.south)!0.5!(review.south) + (0,-0.35)$) {feedback and revisions};
\end{tikzpicture}
\caption{Workflow on the Vavanagi platform, with one admin user, 77 translators, and 4 reviewers.}
\label{fig:vavanagi-workflow}
\end{figure*}

\subsection{Project initiative and governance}

\textbf{Inception.} The Hula community is organised online around a WhatsApp community, within which there are interest groups (e.g., canoe racing, church, history, language). In the language interest group, members discussed long-running Bible translation efforts and whether AI could help accelerate translation work. This raised the need for crowdsourcing a Hula corpus. The initiative was led by a community member with a technical background, who was inspired by related low-resource language efforts \cite{merx-etal-2025-low}.

\textbf{Development process.} Early development started in Replit\footnote{https://replit.com/} and later moved to a code editor as the platform matured. The initial workflow focused on text translation, reflecting the initial Bible translation use case. Voice capture was then added to make participation easier for elders who are less comfortable writing Hula, and to support future speech technologies (e.g., ASR).

\textbf{Dissemination and governance.}
Recruitment followed existing community channels: Vavanagi platform reviewers are the administrators of the WhatsApp language interest group, and translators are members of the group. This creates a natural mapping from existing community structures to platform roles. Once published, uptake accelerated and the project was featured by NBC Radio, the national broadcaster in Papua New Guinea.

\textbf{Community financing.} 
Community financing emerged as part of governance. Community members in urban areas, who may use Hula less in their daily lives, but who have more disposable income, contribute to a shared prize pool that rewards village-based translators who complete most of the translation work. The current incentive is PGK0.10 per sentence, with voluntary contributions of PGK10–PGK100 from non-translating members. This model distributes participation, and reinforces collective ownership of the language, including for those who may use Hula less frequently, but who want to participate in its cultural preservation.

\subsection{Platform design}

\subsubsection{Workflow}

\textbf{Pipeline structure.} The platform has a four-stage pipeline (Figure \ref{fig:vavanagi-workflow}). First, an administrator imports English source material to create translation tasks. Second, community translators submit Hula translations in text and, when appropriate, as voice recordings. Third, a reviewer team evaluates submissions by checking linguistic clarity, cultural appropriateness, and consistency across entries. Finally, the administrator exports approved records for archiving and downstream use. This role separation enables high participation while keeping quality control explicit.

\textbf{Data revision.} The workflow also includes an iterative feedback loop between translators and reviewers. Rather than rejecting entries as final failures, reviewers flag issues and leave comments that guide revision, after which translators or other reviewers can submit improved versions. This process balances speed with reliability and helps the corpus reflect shared norms and variation in Hula usage.

\subsubsection{Features}

\textbf{Contribution features.} Translators can enter text translations, attach or record audio, and view task-level progress as they work through batches of input text. Reviewers can annotate entries, flag problems, and record approval decisions in a structured way. A dashboard gives an overview of translation progress, with a leaderboard that brings some social motivation to the process.

\textbf{Administration features.} Role-based permissions differentiate administration, translation, and review. Import and export functions support batch operations, making it practical to transfer data to and from research projects. These capabilities make the platform a lightweight coordination system for collective resource gathering.

\begin{figure*}[t]
\centering
\begin{minipage}[t]{0.33\linewidth}
    \includegraphics[width=\linewidth]{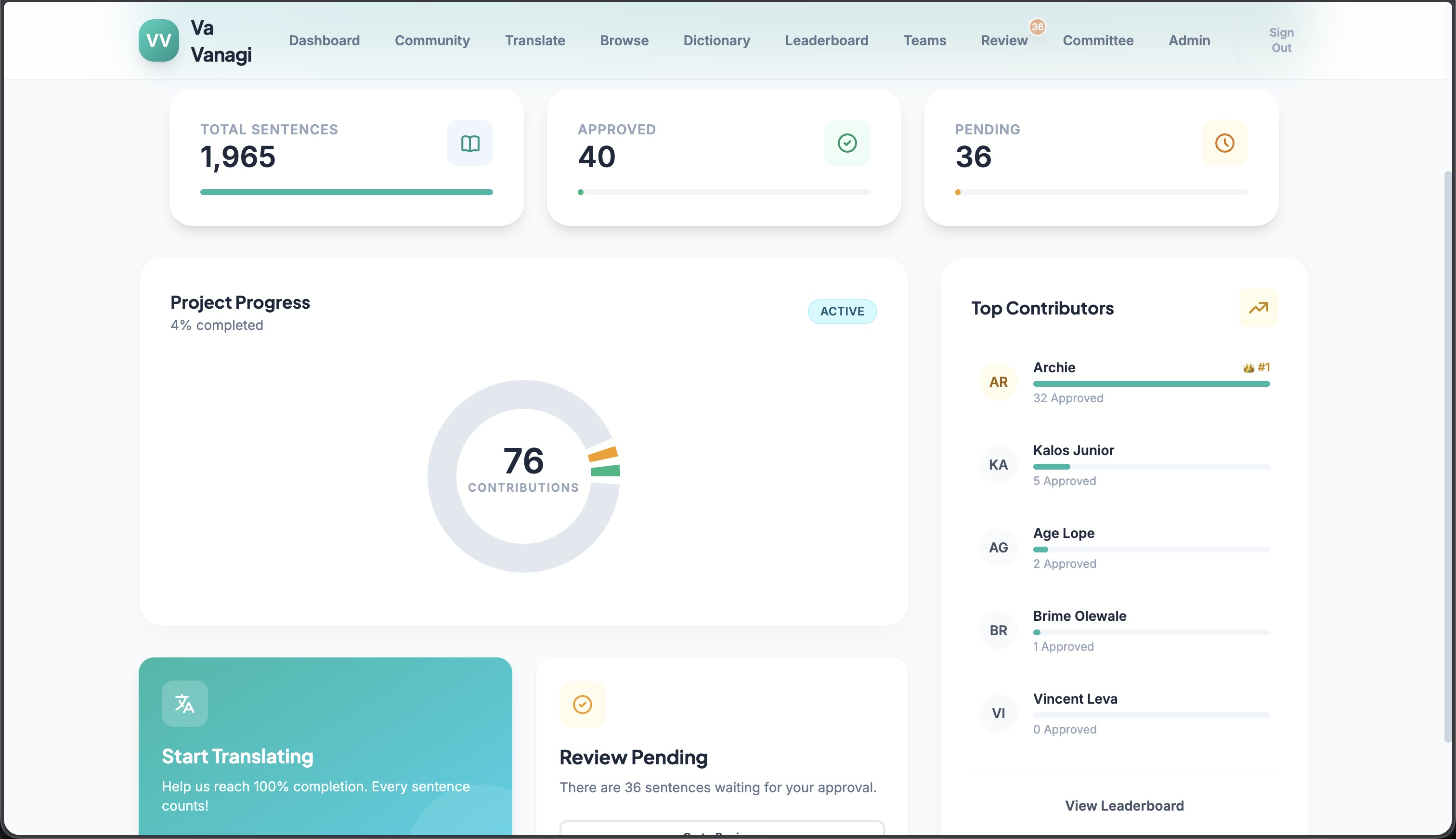}
    \centering\small (a) Home page
\end{minipage}\hfil
\begin{minipage}[t]{0.33\linewidth}
    \includegraphics[width=\linewidth]{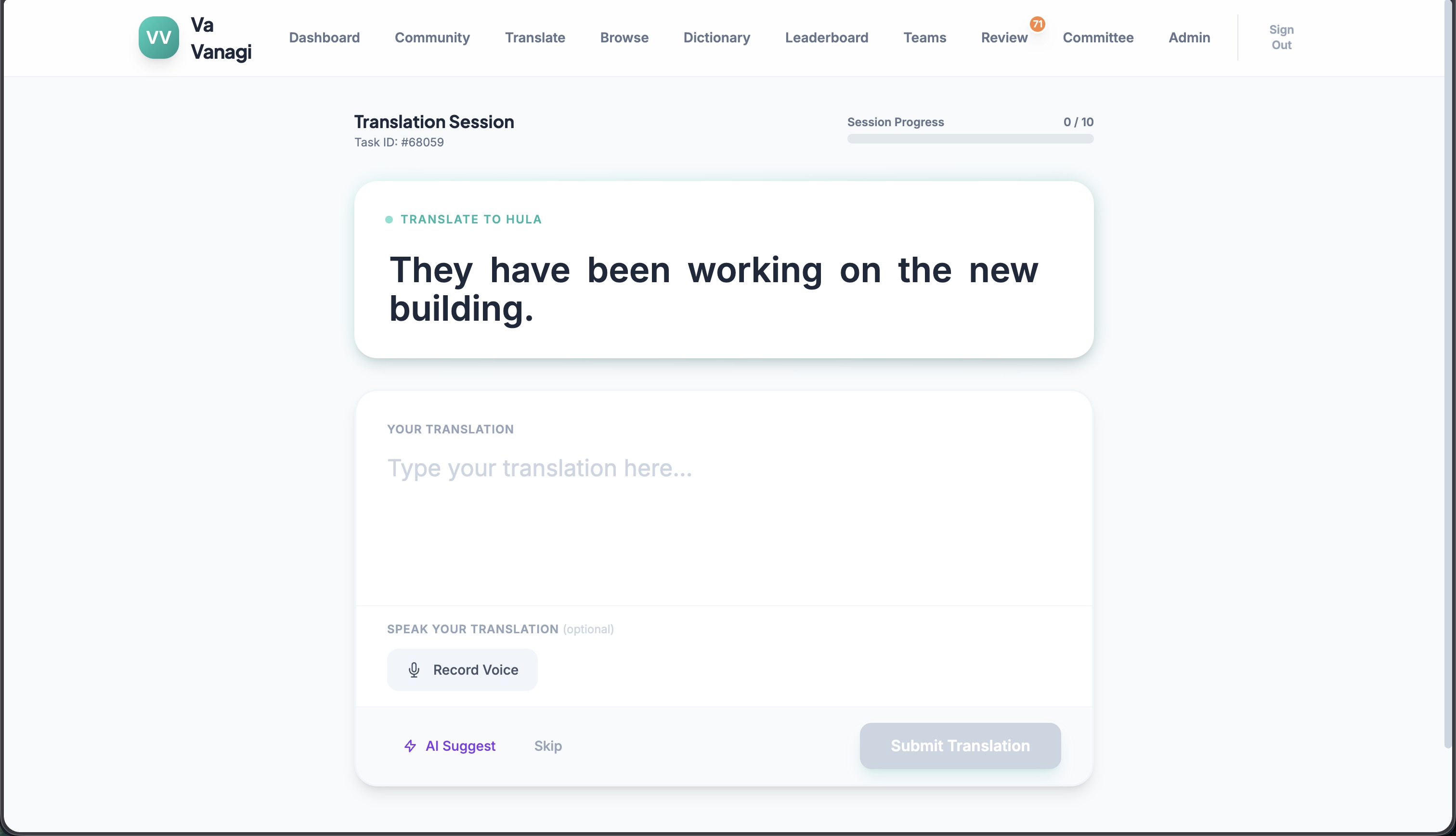}
    \centering\small (b) Data entry
\end{minipage}\hfil
\begin{minipage}[t]{0.33\linewidth}
    \includegraphics[width=\linewidth]{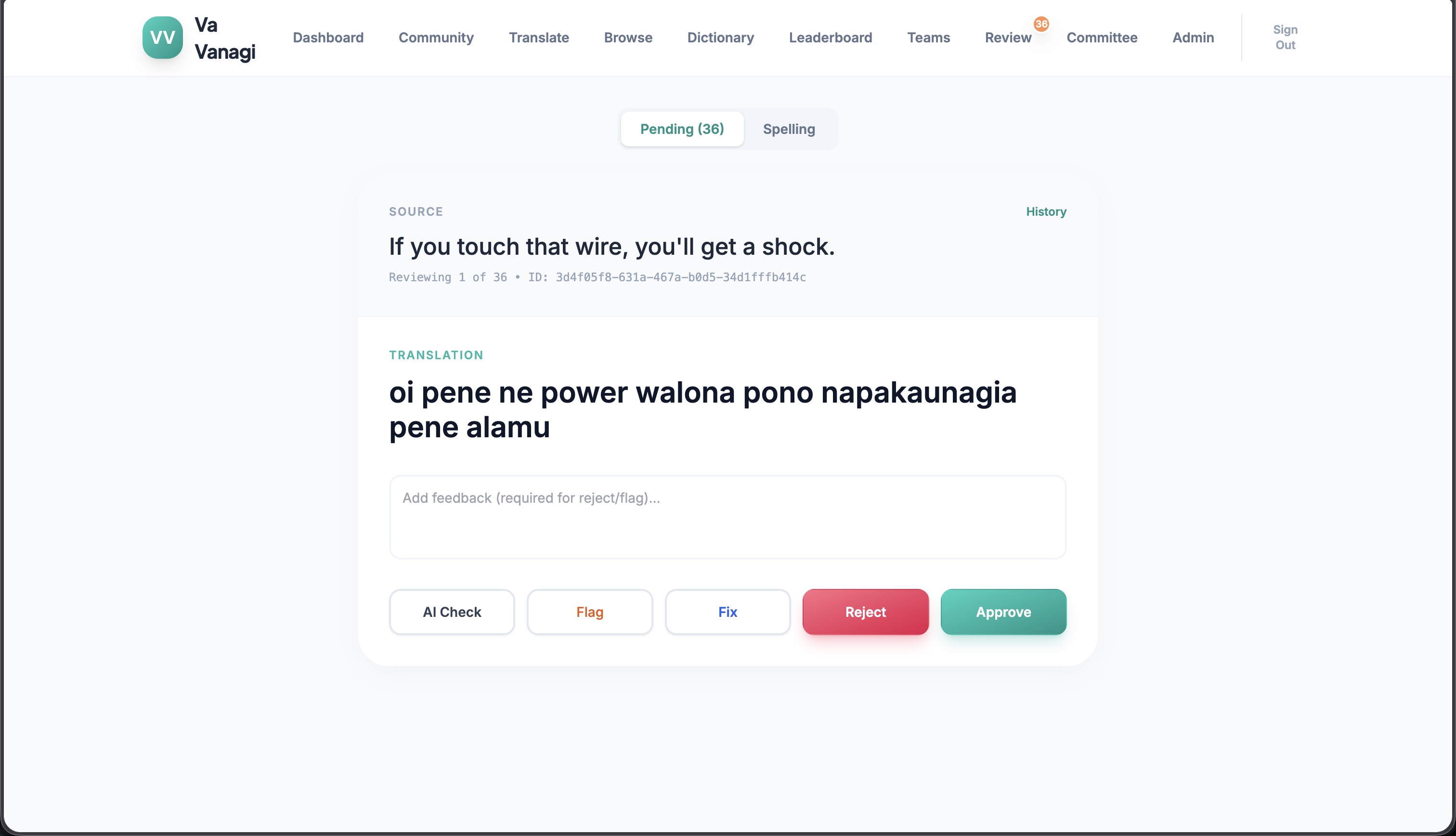}
    \centering\small (c) Review
\end{minipage}
\caption{Screenshots of the Vavanagi platform showing the home page, translator data-entry view, and reviewer interface.}
\label{fig:screenshots}
\end{figure*}

\subsubsection{Data}

\textbf{Database.} Vavanagi stores records in Firebase Firestore, a cloud document database that can be used directly by the application without deploying a separate hosted backend. Firebase also provides authentication, which supports role-based access and user identity in the platform. In addition, the system records user actions so the team can track who edited which data and when. This setup has proved economical to date, with total project tech costs (including domain registration) of less than \$20 USD.

\textbf{Data structure.} A `Sentence` record stores the English source prompt and task-level metadata (e.g., status and assignment state). A `Translation` record stores a contributor's Hula translation of a sentence, together with submission metadata. A `TranslationReview` record stores reviewer decisions and comments on translations. A `User` record stores identity, role, and participation-related information used for access control and activity tracking.

\section{Results}

\subsection{Corpus and community}

\begin{table}[t]
\centering
\small
\caption{Vavanagi corpus and participation statistics.}
\label{tab:corpus-stats}
\begin{tabular}{lr}
\toprule
\textbf{Metric} & \textbf{Value} \\
\midrule
Sentence pairs & 12{,}124+ \\
Unique English words & 7{,}948 \\
Unique Hula words & 9{,}556 \\
Median sentence length & 8 words / 39 characters \\
Approved after 1 / 2 / 3+ translations & 91\% / 8\% / 1\% \\
\midrule
Translators & 77 \\
Reviewers & 4 \\
\bottomrule
\end{tabular}
\end{table}

Table~\ref{tab:corpus-stats} summarises the corpus and participation figures. Uptake was rapid: the initial batch of 2{,}000 sentences was completed in two weeks, and a second batch of 1{,}500 in three days. Sentences are short and accessible (median 8 words), which likely supports both speed and consistency. The high first-pass approval rate (91\%) suggests that reviewers and translators are aligned on translation quality.

Reviewers noted that the translations capture the rise of loanwords from Tok Pisin (e.g., \textit{ketolo} for kettle, \textit{lamepa} for lamp) and of code-switching. They reported this reflects a broader language transformation in the village, where adolescents often use Tok Pisin with one another despite understanding the Hula of their parents, a well-documented pattern that puts smaller languages at risk of being supplanted \cite{kik2021language}.

\subsection{User feedback}

We collected System Usability Scale \cite{brooke1996sus} (SUS) responses from 8 translators, yielding a mean score of 73.4, indicating above-average usability. Translators strongly disagreed that ``the system is unnecessarily complex'' (mean 1.5/5), confirming that Vavanagi is easy to use and fit for purpose, but rated ``I would like to use this system frequently'' low (3.4/5), coherent with a platform designed for a specific task rather than daily use. In open comments, translators emphasised the value of audio for capturing correct Hula pronunciation, noted that the platform could allow entering alternative Hula translations for a single English input, and requested the ability to view similar existing translations while working. This feedback points to concrete directions for future work.

\section{Discussion}

\textbf{Community agency.} 
Rather than positioning the Hula community as a partner in ethical research practice, Vavanagi shows self-determined collective agency: community members designed and institutionalised their own governance mechanisms and digital infrastructure, exercising technological sovereignty over their language.
This differs from ethics framings centred on consultation, harm minimisation, and benefit sharing: while these remain necessary, they do not always result in an initiative where the community feels agency and ownership over the project. Overall, Vavanagi resonates with CARE principles \cite{care-principles} (collective benefit, authority to control, responsibility, and ethics) in ways that match a self-governing, high-agency community position.

\textbf{Language technology as a bridge between generations.}
The Hula language remains strong in the Hula village, but participants describe reduced everyday use among urban migrants and in mixed-language households. At the same time, many community members expressed a strong desire to reconnect with their linguistic heritage. Vavanagi suggests that language technology can mediate this tension by bridging distance (between village-based and urban community members) and generation (between elder speakers and younger learners). Importantly, the platform's interactive design (voice features and collaborative workflows) increases the language's perceived relevance and prestige, positioning Hula as a living language of the present that the community can be proud of.

\textbf{AI tools and avoiding a "one-size-fits-all" approach.}
Vavanagi shows how AI coding tools can support community-specific innovation, when used as enabling infrastructure rather than as fixed solutions. In this case, a community member who is technical but not a web developer leveraged AI coding assistants to tailor the platform to community context. The result is not simply a lower-cost implementation, but a system whose interaction model reflects local governance. This resonates with prior work cautioning against one-size-fits-all language technologies and advocating adaptable ``digital shell'' approaches that communities can customise to their own priorities \cite{cooper-etal-2024-things}.


\section{Future Work}

\textbf{Speech technology development.} 
Building on the parallel text and audio corpus gathered through Vavanagi, our primary technical objective is the development of downstream machine learning models tailored to the Hula language. We plan to train an initial machine translation model, followed by an ASR system. These models will serve as the foundational infrastructure for practical, community-facing tools, such as a voice-enabled Hula-English translation application designed to facilitate everyday market commerce and media consumption. 

\textbf{Pure voice input pipeline.}
The current platform requires translators to enter text, which creates a participation barrier for elders who are less used to writing Hula. A planned extension will allow contributors to submit voice-only responses, with a second contributor transcribing the audio into text. This aims to split the work along capability levels: older speakers are most fluent in Hula, but have less digital literacy, while younger speakers have more digital literacy, but are less fluent in Hula. This approach also bridges the gap between collecting data for translation and for speech recognition.

\textbf{Cultural expansion beyond translation.}
The platform infrastructure could support broader cultural documentation: oral histories, traditional knowledge, place names, and songs. These domains would enrich the linguistic resource while reinforcing the community's relationship to Hula as a living cultural heritage, not only a communication system.




\section{Conclusion}

Vavanagi is a community-led platform for crowdsourced translation and language documentation for Hula. The project has engaged 77 translators and 4 reviewers in producing 12,124 parallel English-Hula sentences.

Through a proposed five-level community involvement spectrum, we position Vavanagi as a Level 5 project: fully community-initiated and governed. To our knowledge, this makes it the first community-led language technology initiative for a language of this size (10k speakers).

Beyond the corpus, Vavanagi illustrates how language technology can serve as a bridge: between village-based members and urban diaspora, between generations, and between cultural heritage and contemporary digital life. The community governance, financing model, and platform architecture reflect locally grounded solutions to challenges that are common across under-resourced Pacific languages.

We hope Vavanagi offers a transferable model for other Pacific language communities seeking to leverage technology for language preservation on their own terms.

\section{Acknowledgements}
We thank the Vula'a Kunenai Community, and in particular the translators and reviewers whose contributions made this corpus possible. This research was supported by an Australian Research Council Discovery Early Career Researcher Award (DECRA project number DE260100695) held by Ekaterina Vylomova.

\section{Use of Generative AI Disclosure}
Generative AI tools were used in two ways during the preparation of this manuscript. First, AI writing assistants were used to improve grammar, phrasing, and the clarity of text drafted by the authors. Second, AI coding assistants were used to help write code for computing corpus and participation statistics. All AI-assisted output was reviewed and verified by the authors, who take full responsibility for the content of this paper. The use of AI coding tools within the Vavanagi platform itself is part of the contribution described in this work and is reported in the body of the paper.

\bibliographystyle{IEEEtran}
\bibliography{citations}

\begin{thebibliography}{10}
\providecommand{\url}[1]{#1}
\csname url@samestyle\endcsname
\providecommand{\newblock}{\relax}
\providecommand{\bibinfo}[2]{#2}
\providecommand{\BIBentrySTDinterwordspacing}{\spaceskip=0pt\relax}
\providecommand{\BIBentryALTinterwordstretchfactor}{4}
\providecommand{\BIBentryALTinterwordspacing}{\spaceskip=\fontdimen2\font plus
\BIBentryALTinterwordstretchfactor\fontdimen3\font minus
  \fontdimen4\font\relax}
\providecommand{\BIBforeignlanguage}[2]{{%
\expandafter\ifx\csname l@#1\endcsname\relax
\typeout{** WARNING: IEEEtran.bst: No hyphenation pattern has been}%
\typeout{** loaded for the language `#1'. Using the pattern for}%
\typeout{** the default language instead.}%
\else
\language=\csname l@#1\endcsname
\fi
#2}}
\providecommand{\BIBdecl}{\relax}
\BIBdecl

\bibitem{foley2000languages}
\BIBentryALTinterwordspacing
W.~A. Foley, ``The languages of {New Guinea},'' \emph{Annual Review of
  Anthropology}, vol.~29, pp. 357--404, 2000. [Online]. Available:
  \url{https://doi.org/10.1146/annurev.anthro.29.1.357}
\BIBentrySTDinterwordspacing

\bibitem{aikhenvaldstebbins2007languages}
A.~Y. Aikhenvald and T.~N. Stebbins, ``Languages of {New Guinea},'' in
  \emph{Vanishing Languages of the Pacific Rim}, O.~Miyaoka, O.~Sakiyama, and
  M.~E. Krauss, Eds.\hskip 1em plus 0.5em minus 0.4em\relax Oxford: Oxford
  University Press, 2007, pp. 239--266.

\bibitem{aikhenvald2014living}
\BIBentryALTinterwordspacing
A.~Y. Aikhenvald, ``Living in many languages: linguistic diversity and
  multilingualism in {Papua New Guinea},'' \emph{Language and Linguistics in
  Melanesia}, vol.~32, no.~2, pp. I--XVII, 2014. [Online]. Available:
  \url{https://researchonline.jcu.edu.au/37334/}
\BIBentrySTDinterwordspacing

\bibitem{ross1988protooceanic}
\BIBentryALTinterwordspacing
M.~Ross, \emph{{Proto-Oceanic} and the {Austronesian} Languages of {Western
  Melanesia}}, ser. Pacific Linguistics: Series C.\hskip 1em plus 0.5em minus
  0.4em\relax Canberra: Research School of Pacific and Asian Studies,
  Australian National University, 1988, vol.~98. [Online]. Available:
  \url{http://hdl.handle.net/11858/00-001M-0000-0012-7B5D-4}
\BIBentrySTDinterwordspacing

\bibitem{dutton1973checklist}
\BIBentryALTinterwordspacing
T.~E. Dutton, \emph{A Checklist of Languages and Present-day Villages of
  Central and South-east Mainland {Papua}}, ser. Pacific Linguistics, Series
  B.\hskip 1em plus 0.5em minus 0.4em\relax Canberra: The Australian National
  University, 1973, vol.~24. [Online]. Available:
  \url{https://sealang.net/archives/pl/pdf/PL-B24.pdf}
\BIBentrySTDinterwordspacing

\bibitem{ethnologuehul}
\BIBentryALTinterwordspacing
Ethnologue, ``{Vula'a} language ({HUL}) - {L1} \& {L2} speakers, status, map,
  endangered level \& official use,'' Ethnologue: Languages of the World, 2026,
  accessed: 2026-02-16. [Online]. Available:
  \url{https://www.ethnologue.com/language/hul/}
\BIBentrySTDinterwordspacing

\bibitem{schreyerwagner2022uncertainty}
\BIBentryALTinterwordspacing
C.~Schreyer and J.~Wagner, ``Uncertainty in diversity: language shift and
  language planning in {Papua New Guinea}, a {Kala} case study,'' \emph{Journal
  of Multilingual and Multicultural Development}, 2022. [Online]. Available:
  \url{https://doi.org/10.1080/01434632.2022.2036744}
\BIBentrySTDinterwordspacing

\bibitem{bowern2015lone}
C.~Bowern and N.~Warner, ``‘lone wolves’ and collaboration: A reply to
  {Crippen} \& {Robinson} (2013),'' 2015.

\bibitem{bax2024milpa}
A.~Bax, M.~Bucholtz, E.~W. Campbell, A.~Z. Fawcett, I.~G. Mendoza, S.~L.
  Peters, and G.~R. Basurto, ``{MILPA}: A community-centered linguistic
  collaboration supporting diasporic {Mexican} {Indigenous} ({Ind{\'\i}gena})
  languages in {California},'' \emph{Language Documentation \& Conservation},
  vol.~18, pp. 148--175, 2024.

\bibitem{krajinovic2022communityled}
A.~Krajinović, R.~Billington, L.~Emil, G.~Kaltapau, and N.~Thieberger,
  ``Community-led documentation of {Nafsan} ({Erakor}, {Vanuatu}),'' in
  \emph{Human Language Technology. Challenges for Computer Science and
  Linguistics}, Z.~Vetulani, P.~Paroubek, and M.~Kubis, Eds.\hskip 1em plus
  0.5em minus 0.4em\relax Springer International Publishing, 2022, pp.
  112--128.

\bibitem{masakhane-mt-emnlp20}
\BIBentryALTinterwordspacing
W.~Nekoto, V.~Marivate, T.~Matsila, T.~Fasubaa, T.~Fagbohungbe, S.~O. Akinola,
  S.~Muhammad, S.~Kabongo~Kabenamualu, S.~Osei, F.~Sackey, R.~A. Niyongabo,
  R.~Macharm, P.~Ogayo, O.~Ahia, M.~M. Berhe, M.~Adeyemi, M.~Mokgesi-Selinga,
  L.~Okegbemi, L.~Martinus, K.~Tajudeen, K.~Degila, K.~Ogueji, K.~Siminyu,
  J.~Kreutzer, J.~Webster, J.~T. Ali, J.~Abbott, I.~Orife, I.~Ezeani, I.~A.
  Dangana, H.~Kamper, H.~Elsahar, G.~Duru, G.~Kioko, M.~Espoir, E.~van Biljon,
  D.~Whitenack, C.~Onyefuluchi, C.~C. Emezue, B.~F.~P. Dossou, B.~Sibanda,
  B.~Bassey, A.~Olabiyi, A.~Ramkilowan, A.~{\"O}ktem, A.~Akinfaderin, and
  A.~Bashir, ``Participatory research for low-resourced machine translation: A
  case study in {A}frican languages,'' in \emph{Findings of the Association for
  Computational Linguistics: EMNLP 2020}, T.~Cohn, Y.~He, and Y.~Liu,
  Eds.\hskip 1em plus 0.5em minus 0.4em\relax Online: Association for
  Computational Linguistics, Nov. 2020, pp. 2144--2160. [Online]. Available:
  \url{https://aclanthology.org/2020.findings-emnlp.195/}
\BIBentrySTDinterwordspacing

\bibitem{adelani-etal-2022-thousand}
\BIBentryALTinterwordspacing
D.~I. Adelani, J.~O. Alabi, A.~Fan, J.~Kreutzer, X.~Shen, M.~Reid, D.~Ruiter,
  D.~Klakow, P.~Nabende, E.~Chang, T.~Gwadabe, F.~Sackey, B.~F.~P. Dossou,
  C.~Emezue, C.~Leong, M.~Beukman, S.~H. Muhammad, G.~D. Jarso, O.~Yousuf,
  A.~N. Niyongabo~Rubungo, G.~Hacheme, E.~P. Wairagala, M.~U. Nasir, B.~A.
  Ajibade, T.~O. Ajayi, Y.~W. Gitau, J.~Abbott, M.~Ahmed, M.~Ochieng, A.~Aremu,
  P.~Ogayo, J.~Mukiibi, F.~Ouoba~Kabore, G.~K. Kalipe, D.~Mbaye, A.~A. Tapo,
  V.~M. Memdjokam~Koagne, E.~Munkoh-Buabeng, V.~Wagner, I.~Abdulmumin,
  A.~Awokoya, H.~Buzaaba, B.~Sibanda, A.~Bukula, and S.~Manthalu, ``A few
  thousand translations go a long way! leveraging pre-trained models for
  {A}frican news translation,'' in \emph{Proceedings of the 2022 Conference of
  the North American Chapter of the Association for Computational Linguistics:
  Human Language Technologies}, M.~Carpuat, M.-C. de~Marneffe, and I.~V.
  Meza~Ruiz, Eds.\hskip 1em plus 0.5em minus 0.4em\relax Seattle, United
  States: Association for Computational Linguistics, Jul. 2022, pp. 3053--3070.
  [Online]. Available: \url{https://aclanthology.org/2022.naacl-main.223/}
\BIBentrySTDinterwordspacing

\bibitem{oneil-etal-2024-computational}
\BIBentryALTinterwordspacing
A.~O{'}Neil, D.~Swanson, and S.~Chelliah, ``Computational language
  documentation: Designing a modular annotation and data management tool for
  cross-cultural applicability,'' in \emph{Proceedings of the 2nd Workshop on
  Cross-Cultural Considerations in NLP}, V.~Prabhakaran, S.~Dev, L.~Benotti,
  D.~Hershcovich, L.~Cabello, Y.~Cao, I.~Adebara, and L.~Zhou, Eds.\hskip 1em
  plus 0.5em minus 0.4em\relax Bangkok, Thailand: Association for Computational
  Linguistics, Aug. 2024, pp. 107--116. [Online]. Available:
  \url{https://aclanthology.org/2024.c3nlp-1.9/}
\BIBentrySTDinterwordspacing

\bibitem{liu-etal-2022-always}
\BIBentryALTinterwordspacing
Z.~Liu, C.~Richardson, R.~Hatcher, and E.~Prud{'}hommeaux, ``Not always about
  you: Prioritizing community needs when developing endangered language
  technology,'' in \emph{Proceedings of the 60th Annual Meeting of the
  Association for Computational Linguistics (Volume 1: Long Papers)},
  S.~Muresan, P.~Nakov, and A.~Villavicencio, Eds.\hskip 1em plus 0.5em minus
  0.4em\relax Dublin, Ireland: Association for Computational Linguistics, May
  2022, pp. 3933--3944. [Online]. Available:
  \url{https://aclanthology.org/2022.acl-long.272/}
\BIBentrySTDinterwordspacing

\bibitem{hutchinson2025designing}
B.~Hutchinson, C.~R. Louro, G.~Collard, and N.~Cooper, ``Designing speech
  technologies for {Australian} {Aboriginal} {English}: Opportunities, risks
  and participation,'' in \emph{Proceedings of the 2025 ACM Conference on
  Fairness, Accountability, and Transparency}, 2025, pp. 108--124.

\bibitem{cooper-etal-2024-things}
\BIBentryALTinterwordspacing
N.~Cooper, C.~Heldreth, and B.~Hutchinson, ````it{'}s how you do things that
  matters'': Attending to process to better serve {Indigenous} communities with
  language technologies,'' in \emph{Proceedings of the 18th Conference of the
  European Chapter of the Association for Computational Linguistics (Volume 2:
  Short Papers)}, Y.~Graham and M.~Purver, Eds.\hskip 1em plus 0.5em minus
  0.4em\relax St. Julian{'}s, Malta: Association for Computational Linguistics,
  Mar. 2024, pp. 204--211. [Online]. Available:
  \url{https://aclanthology.org/2024.eacl-short.19/}
\BIBentrySTDinterwordspacing

\bibitem{mdc2025}
{Mozilla Foundation}, ``Mozilla data collective launches,''
  \url{https://www.mozillafoundation.org/en/meet-mozilla/press-center/mozilla-data-collective-launches/},
  2025, accessed: 2026-02-15.

\bibitem{watimelu2024community}
W.~O.~S. Watimelu, ``Community-led initiatives in language preservation:
  Strategies for endangered language documentation and revitalization,''
  \emph{LIER: Language Inquiry \& Exploration Review}, vol.~1, no.~1, pp. 1--9,
  2024.

\bibitem{cox-etal-2025-creating}
\BIBentryALTinterwordspacing
C.~Cox, B.~Starlight, J.~Crane-Starlight, H.~Big~Crow, and A.~Arppe, ``Creating
  an intelligent dictionary of {Tsuut{'}ina} one verb at a time,'' in
  \emph{Proceedings of the Eight Workshop on the Use of Computational Methods
  in the Study of Endangered Languages}, J.~Lachler, G.~Agyapong, A.~Arppe,
  S.~Moeller, A.~Chaudhary, S.~Rijhwani, and D.~Rosenblum, Eds.\hskip 1em plus
  0.5em minus 0.4em\relax Honolulu, Hawaii, USA: Association for Computational
  Linguistics, Mar. 2025, pp. 110--119. [Online]. Available:
  \url{https://aclanthology.org/2025.computel-main.12/}
\BIBentrySTDinterwordspacing

\bibitem{mirzakhalov2021large}
J.~Mirzakhalov, A.~Babu, D.~Ataman, S.~Kariev, F.~Tyers, O.~Abduraufov,
  M.~Hajili, S.~Ivanova, A.~Khaytbaev, A.~Laverghetta~Jr \emph{et~al.}, ``A
  large-scale study of machine translation in {Turkic} languages,'' in
  \emph{Proceedings of the 2021 Conference on Empirical Methods in Natural
  Language Processing}, 2021, pp. 5876--5890.

\bibitem{hameed25_interspeech}
R.~Hameed, S.~Ahmadi, H.~Hadi, and R.~Sennrich, ``{Automatic Speech Recognition
  for Low-Resourced Middle Eastern Languages},'' in \emph{{Interspeech 2025}},
  2025, pp. 733--737.

\bibitem{petrariu-nisioi-2024-multilingual}
\BIBentryALTinterwordspacing
I.~Petrariu and S.~Nisioi, ``A multilingual parallel corpus for {A}romanian,''
  in \emph{Proceedings of the 2024 Joint International Conference on
  Computational Linguistics, Language Resources and Evaluation (LREC-COLING
  2024)}, N.~Calzolari, M.-Y. Kan, V.~Hoste, A.~Lenci, S.~Sakti, and N.~Xue,
  Eds.\hskip 1em plus 0.5em minus 0.4em\relax Torino, Italia: ELRA and ICCL,
  May 2024, pp. 832--838. [Online]. Available:
  \url{https://aclanthology.org/2024.lrec-main.75/}
\BIBentrySTDinterwordspacing

\bibitem{li-etal-2024-speech}
\BIBentryALTinterwordspacing
Z.~Li, M.~Rind-Pawlowski, and J.~Niehues, ``Speech recognition corpus of the
  {Khinalug} language for documenting endangered languages,'' in
  \emph{Proceedings of the 2024 Joint International Conference on Computational
  Linguistics, Language Resources and Evaluation (LREC-COLING 2024)},
  N.~Calzolari, M.-Y. Kan, V.~Hoste, A.~Lenci, S.~Sakti, and N.~Xue, Eds.\hskip
  1em plus 0.5em minus 0.4em\relax Torino, Italia: ELRA and ICCL, May 2024, pp.
  15\,171--15\,180. [Online]. Available:
  \url{https://aclanthology.org/2024.lrec-main.1319/}
\BIBentrySTDinterwordspacing

\bibitem{prabowo2024preserving}
Y.~D. Prabowo, M.~Gabriel, R.~Nazarudin, T.~G. Ratumanan, and M.~Maslim,
  ``Preserving {Meher} and {Woirata} corpus languages using neural machine
  translation,'' \emph{Indonesian Journal of Information Systems}, vol.~6,
  no.~2, pp. 156--175, 2024.

\bibitem{frontull-etal-2025-bringing}
\BIBentryALTinterwordspacing
S.~Frontull, T.~Str{\"o}hle, C.~Zoli, W.~Pescosta, U.~Frenademez, M.~Ruggeri,
  D.~Valentin, K.~Comploj, G.~Perathoner, S.~Liotto, and P.~Anvidalfarei,
  ``Bringing {L}adin to {FLORES}+,'' in \emph{Proceedings of the Tenth
  Conference on Machine Translation}, B.~Haddow, T.~Kocmi, P.~Koehn, and
  C.~Monz, Eds.\hskip 1em plus 0.5em minus 0.4em\relax Suzhou, China:
  Association for Computational Linguistics, Nov. 2025, pp. 1061--1071.
  [Online]. Available: \url{https://aclanthology.org/2025.wmt-1.81/}
\BIBentrySTDinterwordspacing

\bibitem{merx-etal-2025-low}
\BIBentryALTinterwordspacing
R.~Merx, A.~J.~G. Correia, H.~Suominen, and E.~Vylomova, ``Low-resource machine
  translation: what for? who for? an observational study on a dedicated {Tetun}
  language translation service,'' in \emph{Proceedings of the Eighth Workshop
  on Technologies for Machine Translation of Low-Resource Languages (LoResMT
  2025)}, A.~K. Ojha, C.-h. Liu, E.~Vylomova, F.~Pirinen, J.~Washington,
  N.~Oco, and X.~Zhao, Eds.\hskip 1em plus 0.5em minus 0.4em\relax Albuquerque,
  New Mexico, U.S.A.: Association for Computational Linguistics, May 2025, pp.
  54--65. [Online]. Available: \url{https://aclanthology.org/2025.loresmt-1.7/}
\BIBentrySTDinterwordspacing

\bibitem{kik2021language}
\BIBentryALTinterwordspacing
A.~Kik, M.~Adamec, A.~Y. Aikhenvald, J.~Bajzekova, N.~Baro, C.~Bowern, R.~K.
  Colwell, P.~Drozd, P.~Duda, S.~Ibalim, L.~R. Jorge, J.~Mogina, B.~Ruli,
  K.~Sam, H.~Sarvasy, S.~Saulei, G.~D. Weiblen, J.~Zrzavy, and V.~Novotny,
  ``Language and ethnobiological skills decline precipitously in {Papua New
  Guinea}, the world’s most linguistically diverse nation,''
  \emph{Proceedings of the National Academy of Sciences}, vol. 118, no.~22, p.
  e2100096118, 2021. [Online]. Available:
  \url{https://www.pnas.org/doi/abs/10.1073/pnas.2100096118}
\BIBentrySTDinterwordspacing

\bibitem{brooke1996sus}
J.~Brooke, ``{SUS}: A `quick and dirty' usability scale,'' in \emph{Usability
  Evaluation in Industry}, P.~W. Jordan, B.~Thomas, B.~A. Weerdmeester, and
  I.~L. McClelland, Eds.\hskip 1em plus 0.5em minus 0.4em\relax London: Taylor
  \& Francis, 1996, pp. 189--194.

\bibitem{care-principles}
\BIBentryALTinterwordspacing
S.~R. Carroll, I.~Garba, O.~L. Figueroa-Rodr{\'\i}guez, J.~Holbrook, R.~Lovett,
  S.~Materechera, M.~Parsons, K.~Raseroka, D.~Rodriguez-Lonebear, R.~Rowe
  \emph{et~al.}, ``The {CARE} principles for {Indigenous} data governance,''
  \emph{Data Science Journal}, vol.~19, no.~1, p.~43, 2020. [Online].
  Available: \url{https://datascience.codata.org/articles/10.5334/dsj-2020-043}
\BIBentrySTDinterwordspacing

\end{thebibliography}

\end{document}